\documentclass[sigconf]{acmart}
\AtBeginDocument{%
  }

\setcopyright{acmlicensed}
\copyrightyear{2026}
\acmYear{2026}
\acmDOI{XXXXXXX.XXXXXXX}
\acmConference[ACM MM’26]{Make sure to enter the correct
  conference title from your rights confirmation email}{June 03--05,
  2026}{Rio de Janeiro, Brazil}
\acmISBN{978-1-4503-XXXX-X/2018/06}

\usepackage{subcaption}

\usepackage{amsmath}
\usepackage{algorithmic}
\usepackage[ruled,vlined,linesnumbered]{algorithm2e}
\usepackage[most]{tcolorbox}
\usepackage{listings}
\usepackage{xcolor}
\usepackage{fancyhdr}
\tcbuselibrary{breakable,listings,skins}

\renewcommand \footnotetextcopyrightpermission[1]{}

\newtcolorbox{promptbox}[2][]{
    colback=gray!3,
    colframe=black!60,
    coltitle=black,
    fonttitle=\bfseries,
    title=#2,
    boxrule=0.6pt,
    arc=2mm,
    left=2mm,
    right=2mm,
    top=1mm,
    bottom=1mm,
    enhanced,
    fontupper=\ttfamily\small,
    breakable,
    extras first={
        after upper={\vfill}, 
        bottomrule=0.6pt,
    },
    extras middle={
        toprule=0.6pt,
        bottomrule=0.6pt,
    },
    extras last={
        toprule=0.6pt,
    },
    #1
}

\renewcommand\footnotetextcopyrightpermission[1]{}

\begin{document}

\title{TED: Training-Free Experience Distillation for Multimodal Reasoning}


\author{Shuozhi Yuan}
\authornotemark[1]
\author{Jinqing Wang}
\author{Zihao Liu}
\email{yuansz@chinatelecom.cn}
\affiliation{%
  \institution{China Telecom Digital Intelligence Technology Co.,Ltd.}
  \city{Beijing}
  \country{China}
}



\author{Miaomiao Yuan}
\affiliation{%
  \institution{Institute of Computing Technology, Chinese Academy of Sciences}
  \city{Beijing}
  \country{China}
}

\author{Haoran Peng}
\author{Jin Zhao}
\author{Bingwen Wang}
\author{Haoyi Wang}
\affiliation{%
  \institution{China Telecom Digital Intelligence Technology Co.,Ltd.}
  \city{Beijing}
  \country{China}
}




\renewcommand{\shortauthors}{Shuozhi Yuan,et al.}

\begin{abstract}
Knowledge distillation (KD) is typically realized by transferring a teacher model’s knowledge into a student’s parameters through supervised or reinforcement-based optimization. While effective, such approaches require repeated parameter updates and large-scale training data, limiting their applicability in resource-constrained environments. In this work, we propose TED, a training-free, context-based distillation framework that shifts the update target of distillation from model parameters to an in-context experience injected into the student’s prompt. For each input, the student generates multiple reasoning trajectories, while a teacher independently produces its own solution. The teacher then compares the student trajectories with its reasoning and the ground-truth answer, extracting generalized experiences that capture effective reasoning patterns. These experiences are continuously refined and updated over time. A key challenge of context-based distillation is unbounded experience growth and noise accumulation. TED addresses this with an experience compression mechanism that tracks usage statistics and selectively merges, rewrites, or removes low-utility experiences. Experiments on multimodal reasoning benchmarks MathVision and VisualPuzzles show that TED consistently improves performance. On MathVision, TED raises the performance of Qwen3-VL-8B from 0.627 to \textbf{0.702}, and on VisualPuzzles from 0.517 to \textbf{0.561} with just 100 training samples. Under this low-data, no-update setting, TED achieves performance competitive with fully trained parameter-based distillation while reducing training cost by over 20×, demonstrating that meaningful knowledge transfer can be achieved through contextual experience.
\end{abstract}




\maketitle

\section{Introduction}

\begin{figure}[h]
  \centering
  \includegraphics[width=3.3in]{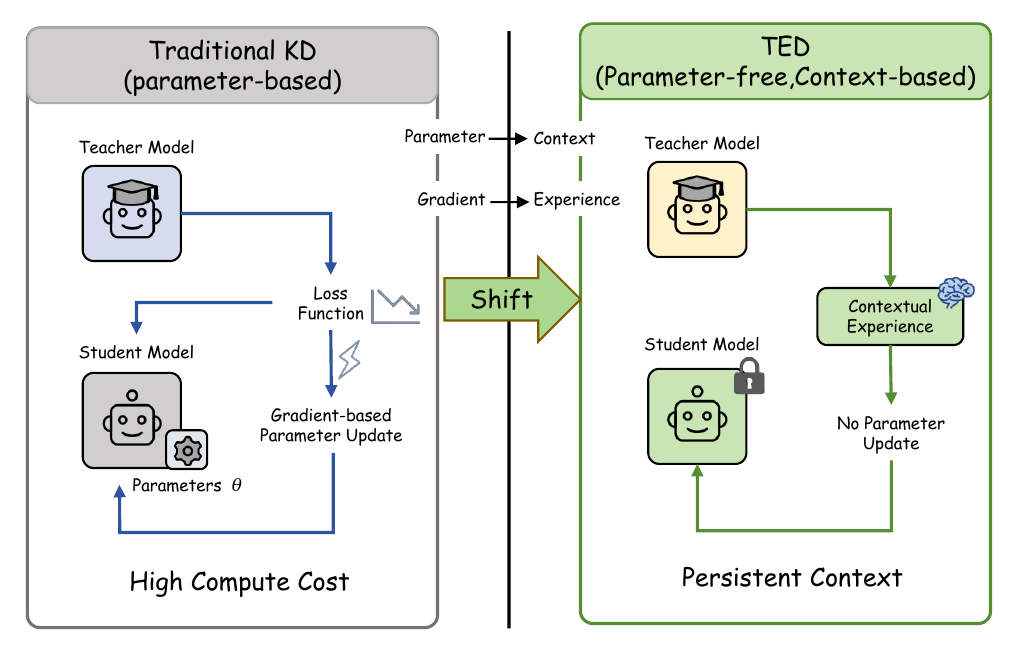}
  \caption{TED reformulates knowledge distillation from parameter updates to contextual experience reuse.}
  \label{Fig1}
  \vspace{-0.5em}
\end{figure}

Knowledge distillation (KD) has become a standard approach for transferring capabilities from multimodal large language models (MLLMs) to smaller ones \cite{Distilling, llava}. Most existing knowledge distillation methods adopt a parameter-based strategy, where the student learns by fine-tuning on large-scale data generated by a teacher, such as soft labels \cite{alpaca}, rationales \cite{Nescd}, or reasoning trajectories \cite{deepseek}. Although effective, these approaches usually rely on gradient-based optimization and repeated parameter updates, which require substantial computational cost \cite{self-instruct, sharegpt4v} and large amounts of training data, limiting their practicality in many resource-constrained or rapidly evolving environments.

For real-world use, especially on edge devices \cite{fang} or black-box APIs \cite{model_distillation}, updating model parameters is impractical or even impossible. In such cases, efficient adaptation without retraining becomes crucial. This raises an important question: \textbf{Can knowledge distillation be achieved without updating model parameters?}

In this work, we answer this question through \textbf{TED}, an alternative formulation of distillation that operates entirely in the model’s context rather than its parameters. As illustrated in Figure 1, unlike traditional distillation methods that encode teacher knowledge into student parameters through optimization, TED reformulates distillation as the continual extraction, abstraction, and reuse of transferable reasoning experiences. These experiences serve as distilled knowledge that guides future inference, enabling the student to improve without any parameter updates.

A key distinction between TED and existing memory-based  methods is what gets stored and updated. Prior methods such as Reflexion\cite{reflexion} and Memento\cite{Memento} typically reuse instance-level trajectories, demonstrations, or verbal feedback from previous attempts. In contrast, TED does not treat experience as a cache of past solutions. Instead, it uses teacher supervision to extract abstract, reusable reasoning experiences—such as transferable strategies, common failure patterns, and correction rules—that generalize across inputs. In this sense, TED distills not example-specific traces, but higher-level reasoning guidance.

For each input, the student model generates multiple reasoning trajectories in parallel, while a teacher model independently produces its own reasoning process. The teacher then jointly evaluates the student trajectories, its own reasoning, and the ground-truth, and abstracts generalized experiences that capture effective reasoning patterns, common failure modes, and correction strategies. Crucially, these experiences are not task-specific exemplars or raw demonstrations, but compact and reusable reasoning principles distilled under teacher supervision. These experiences are accumulated across training samples and iteratively refined, forming a persistent experience that evolves over time. During inference, the learned experience is directly injected into the system prompt, allowing the student model to benefit from distilled knowledge without any parameter updates. This formulation enables TED to operate under strict constraints on training cost and model scale, while remaining compatible with standard black-box APIs.

A key challenge in context-based distillation is that naively accumulating experiences leads to unbounded context growth and low-utility information. TED addresses this challenge through a teacher-guided experience compression mechanism that explicitly models experience utility. Instead of performing simple summarization or heuristic pruning, TED tracks the usage frequency of individual experience items and retains high-utility experiences while selectively merging, rewriting, or removing others under teacher supervision. This compression process abstracts higher-level reasoning patterns from frequently co-occurring experiences and eliminates obsolete or noisy information, ensuring that the in-context experience remains compact, informative, and scalable over long training iterations.

We evaluate TED on multimodal mathematical reasoning and logic benchmarks \cite{mathverse, mathvision, visualpuzzles, logicvista,aime24, aime25} using open-source vision-language models \cite{Qwen3-VL, kimi}. Despite performing no parameter updates and using only 100 training samples, TED achieves substantial performance improvements over direct inference. In particular, TED provides a strong performance-cost trade-off in low-data settings, approaching the gains of conventional parameter-based distillation while reducing training cost by more than 20×. These results demonstrate that effective knowledge transfer can be realized through contextual experience, offering a lightweight and practical alternative to traditional parameter-based distillation.

Our main contributions in this paper are as follows:

\begin{itemize}

\item We propose TED, a training-free, context-based knowledge distillation framework that enables effective knowledge transfer without any parameter updates.

\item TED introduces a teacher-guided experience generation and compression mechanism that distills reusable reasoning principles and maintains a compact, high-utility in-context experience.

\item Experiments on multimodal and textual reasoning benchmarks show that TED substantially improves model performance in low-data settings. Using only 100 training samples, TED achieves performance competitive with conventional distillation while reducing training cost by more than 20×.

\end{itemize}

\section{Related Work}
In this section, we provide an overview of related work on knowledge distillation, in-context learning and other training-free distillation approaches, highlighting their relevance and main differences to our proposed method.

\subsection{Knowledge Distillation}

Knowledge distillation, as pioneered by \cite{Distilling}, has become a basic approach for transferring the capability from a large teacher model into a compact student model. 

In the area of large language models, numerous approaches are proposed to address the task. For instance, instruction-based approaches like Self-Instruct\cite{self-instruct} and Alpaca \cite{alpaca} use teacher models to generate large-scale synthetic datasets for student fine-tuning. Furthermore, reasoning-based methods like Distilling Step-by-Step \cite{Distilling-step-by-step}  extract rationales from the teacher to guide the student’s learning process. Recent reasoning-focused frameworks, such as Beyond Answers \cite{Beyond} and NesyCD\cite{Nescd} , have further improved the quality of rationales by incorporating multi-teacher feedback or symbolic knowledge. Additionally, methods like DeepSeek-R1\cite{deepseek} demonstrate that distilling long-chain reasoning patterns can significantly boost the performance of smaller open-source models.

From the analysis of existing knowledge distillation methods, it becomes clear that most approaches transfer knowledge by updating student model parameters through large-scale optimization, relying on extensive training data and computational resources. Such parameter-centric designs limit their applicability in black-box, resource-constrained, or rapidly evolving settings.
In contrast, our core motivation is to enable effective knowledge distillation without any parameter updates. To this end, we propose TED, which shifts distillation from parameter optimization to context-level experience accumulation, allowing knowledge to be distilled, compressed, and reused entirely through prompts.

\subsection{Multimodal Knowledge Distillation}

With the rise of multimodal large language models (MLLMs), knowledge distillation has extended from pure text to cross-modal reasoning. Early works like MiniGPT-4 \cite{Mini} and LLaVA \cite{llava} focus on aligning visual features with language spaces using teacher-generated data. Recently, more advanced methods attempt to distill specialized capabilities. For instance, Vigstandard \cite{vigc} explores distilling visual grounding capabilities. Similar to the challenges in text-only distillation, MLLM distillation often faces high computational costs due to the large scale of vision-language projectors and encoders. Recent efforts like ShareGPT4V \cite{sharegpt4v} emphasize the quality of teacher-generated captions and rationales to improve student performance with less data. However, most of these MLLM distillation methods still rely on fine-tuning parameters. Our work, TED, offers a potential paradigm shift by demonstrating that multimodal reasoning experiences could also be distilled and accumulated at the context level, potentially bypassing the need for expensive cross-modal fine-tuning.

\subsection{In-context Learning}
In-context learning allows models to perform new tasks by providing a few additional inputs in the prompt without updating any parameters\cite{few_shot}. To improve reasoning performance, Chain-of-Thought \cite{COT} and retrieval-based methods like ERP \cite{ERP} have been proposed to provide reasoning steps or relevant demonstrations. Recent advances, such as Reflexion \cite{reflexion}  and Long-context ICL \cite{long-context}, further use iterative feedback or larger context windows to improve performance.

These studies suggest that in-context learning is a practical way to adapt models without parameter updates. However, most existing methods focus on improving prompt design for individual inputs, such as selecting demonstrations or adding feedback, and do not explicitly model how knowledge can be accumulated across examples. In contrast, TED builds on in-context learning but focuses on experience accumulation. By distilling and reusing reasoning experiences under teacher guidance, TED enables knowledge transfer across inputs in a parameter-free manner.

\subsection{Training-free Distillation}
To further reduce the cost of knowledge transfer, recent research has explored training-free distillation and memory-based learning. For instance, AHA\cite{AHA} and AgentDistill \cite{AgentDistill} facilitate knowledge transfer by collecting successful trajectories and reusing them in the inference stage without any weight updates. More recently, Memento \cite{Memento} introduces a paradigm that allows models to "learn from experience" by storing past successes and failures in an external mental filing cabinet. To address the efficiency issues of large memories, methods like MemCom \cite{Memcom} have been proposed to compress many-shot demonstrations into compact representations.

However, many training-free distillation and memory-based methods mainly reuse raw examples or fixed past trajectories at inference time. When knowledge is treated as a set of instance-level demonstrations, the memory can become noisy and does not generalize well across different inputs. In contrast, TED runs an on-policy-like distillation loop: for each training input, the student samples multiple reasoning trajectories, and the teacher scores them based on the teacher’s own solution and the ground-truth supervision. This trajectory-level comparison allows TED to extract and store abstract experiences (i.e., reusable reasoning tips and common failure patterns), instead of keeping example-specific traces. As a result, TED maintains a compact experience that is continuously updated and is more robust than standard many-shot retrieval when facing noisy or hard cases.

\begin{figure*}[htbp]
  \begin{center}
  \includegraphics[width= 6.8in]{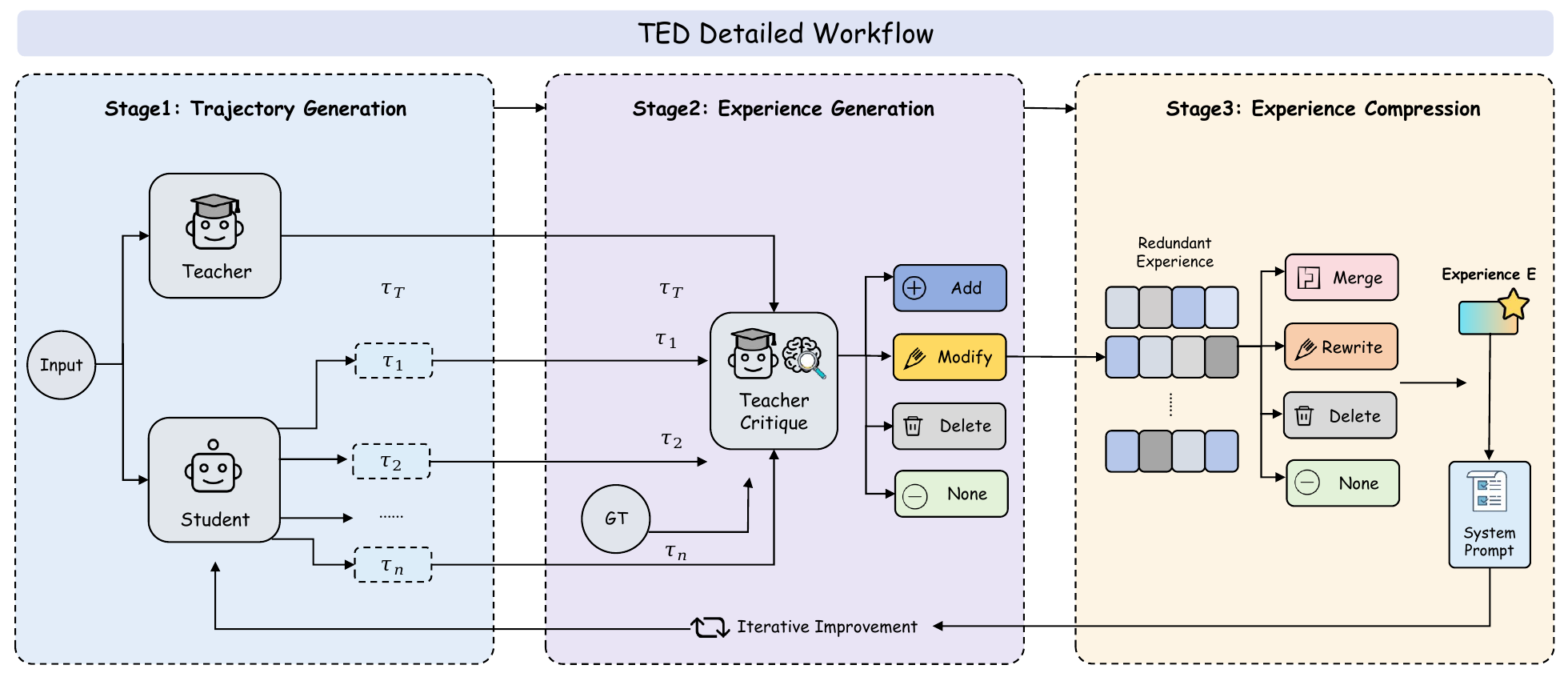}
\caption{Overview of TED. Our proposed method includes three stages: trajectory generation, experience generation, and experience compression. The student first samples multiple reasoning trajectories, which the teacher critiques against its own reasoning and the ground truth to distill generalized experiences. These experiences are then compressed and injected into the system prompt for iterative, parameter-free improvement.}
  \label{Fig2}
  \end{center}
\end{figure*}

\section{Formulation}
We formulate knowledge distillation with a shared on-policy sampling and teacher judging protocol\cite{On-policy}. The difference lies in the update target: vanilla KD updates parameters, while TED updates an in-context experience.

\subsection{On-policy distillation protocol}

Considering supervised examples $(x,y)$, where $x$ is the input and $y$ is the ground-truth.
Let $S$ denote the student model and $T$ the teacher model.
For each input $x$, the student samples $K$ reasoning trajectories
$\{\tau_i\}_{i=1}^{K}$, where each trajectory $\tau$ contains intermediate reasoning and a final answer $\hat{y}(\tau)$.
Independently, the teacher generates its own trajectory $\tau_T$.
A teacher judging module then produces trajectory-level feedback which may incorporate correctness w.r.t.\ $y$ and teacher preference.
\begin{equation}
r_i = \textsc{Judge}\big(\tau_i,\tau_T,y\big),
\end{equation}
This sampling-and-judging step is shared by both vanilla KD and our parameter-free approach.

\subsection{Vanilla KD} 

In vanilla knowledge distillation, the student has trainable parameters $\theta$, and the teacher feedback is converted into a learning signal to optimize $\theta$.
A generic on-policy KD objective can be written as
\begin{equation}
\min_{\theta}\ \mathbb{E}_{(x,y)\sim\mathcal{D}}\ \mathbb{E}_{\tau\sim S_{\theta}(\cdot\mid x)}
\Big[\mathcal{L}_{\text{KD}}\big(\theta;\ x,\tau,\tau_T,y\big)\Big],
\end{equation}
where $\mathcal{L}_{\text{KD}}$ denotes the distillation objective, which can be instantiated as maximizing the likelihood of the best-scored trajectory, preference-based ranking, or reinforcement-style objectives using $r(\tau)$.Training proceeds via repeated gradient updates can be described as:
\begin{equation}
\theta \leftarrow \theta - \eta \nabla_{\theta}\mathcal{L}_{\text{KD}}.
\end{equation}
While effective, this approach requires parameter updates and typically large-scale optimization.

\subsection{TED} TED freezes model parameters and performs distillation by updating a contextual experience  $E$ that is injected into the student's prompt.
We denote the prompted context by
\begin{equation}
p(x;E) = [p_{\text{sys}};\ E;\ x],
\end{equation}
where $p_{\text{sys}}$ is a fixed system instruction and $E$ are experience items in the textual prefix.
The student then samples on-policy trajectories conditioned on this context:
\begin{equation}
\tau_i \sim S(\cdot\mid p(x;E)).
\end{equation}
Instead of optimizing $\theta$, TED updates the experience using teacher feedback:
\begin{equation}
E \leftarrow \textsc{Update}\big(E;\ x,y,\{\tau_i\}_{i=1}^{K},\tau_T,\{r_i\}_{i=1}^{K}\big),
\end{equation}
where \textsc{Update} extracts generalized experience items (reusable reasoning tips and common failure modes) from the comparison among student trajectories, the teacher trajectory, and the ground-truth, and incorporates them into $E$. Specifically, \textsc{Update} is realized as a set of actions generated by teacher model.

At inference time, experience transfer is achieved purely through prompting:
\begin{equation}
\left\{
\begin{aligned}
\tau &\sim S(\cdot\mid p(x;E)), \\
\hat{y} &= \hat{y}(\tau)
\end{aligned}
\right.
\end{equation}

\subsection{Core difference}
Despite sharing the same on-policy sampling and teacher-judging protocol, the two approaches differ in the optimization variable. Vanilla KD updates student parameters via gradient-based optimization, whereas TED freezes parameters and updates a persistent in-context experience serialized into the prompt for subsequent rollouts and inference.
\begin{equation}
\left\{
\begin{aligned}
\text{ KD: } & 
\theta \leftarrow \theta - \eta\nabla_\theta \mathcal{L} \\
\text{TED: } & 
E \leftarrow \textsc{Update}(E;\cdot)
\end{aligned}
\right.
\end{equation}

\section{TED Framework}
As shown in Figure 2, the TED framework consists of three key steps: reasoning trajectory generation, experience generation, and experience compression. Reasoning trajectory generation allows both the student and teacher models to generate their respective reasoning trajectories. Experience generation creates abstract, reusable experience templates based on the teacher model's reasoning path, the student's multiple reasoning paths, and the ground truth. Experience compression further compresses and refines these experiences to prevent context explosion and excessive noise introduction. A detailed explanation of each step is provided in the subsequent section.

\subsection{Reasoning Trajectory Generation}

Given an input--label pair $(x,y)$, TED performs on-policy reasoning trajectory generation for both the student and the teacher models.
The student model $S$ samples $N$ reasoning trajectories in parallel:
\begin{equation}
\{\tilde{\tau}_i\}_{i=1}^{N} \sim S(\cdot \mid p(x;E)),
\end{equation}
where each raw trajectory $\tilde{\tau}$ contains intermediate reasoning traces and a final answer.
In parallel, the teacher model $T$ generates its own raw reasoning trajectory:
\begin{equation}
\tilde{\tau}_T \sim T(\cdot \mid x).
\end{equation}

\subsubsection{Trajectory compression}

Raw reasoning traces from the student and teacher models often contain redundant or noisy content, such as verbose explanations, self-corrections, or exploratory detours.
To make the reasoning paths more concise and reusable, we apply a self-condensation step to each trajectory.

Specifically, for any $\tilde{\tau}$, we ask the same model to rewrite the reasoning into a concise trajectory through prompt engineering.

\begin{equation}
\tau = \textsc{Condense}(\tilde{\tau}),
\end{equation}
where the condensed trajectory $\tau$ is required to follow a same structured format:
\[
\text{Premises} \;\rightarrow\; \text{Step 1} \;\rightarrow\; \text{Step 2} \;\rightarrow\; \cdots \;\rightarrow\; \text{Conclusion}.
\]
This process removes unnecessary content while keeping the key reasoning steps that lead to the final answer.
We apply this step to all student trajectories ${\tilde{\tau}i}{i=1}^{N}$ and the teacher trajectory $\tilde{\tau}T$, resulting in ${\tau_i}{i=1}^{N}$ and $\tau_T$.






\subsubsection{Teacher trajectory filtering.}

To ensure reliable teacher guidance, we only retain teacher trajectories that correctly derive the ground-truth answer. 
Formally, a teacher trajectory $\tau_T$ is considered valid if
\begin{equation}
\hat{y}(\tau_T) = y.
\end{equation}

For samples where the teacher fails to produce a correct reasoning trajectory, we treat them as negative cases and use them to construct critique experience.

Through parallel student sampling, structured trajectory condensation, and teacher filtering, TED produces a set of clean and comparable reasoning trajectories. 
These trajectories serve as the foundation for subsequent experience generation and compression.

\begin{figure*}[htbp]
  \begin{center}
  \includegraphics[width= 7.0in]{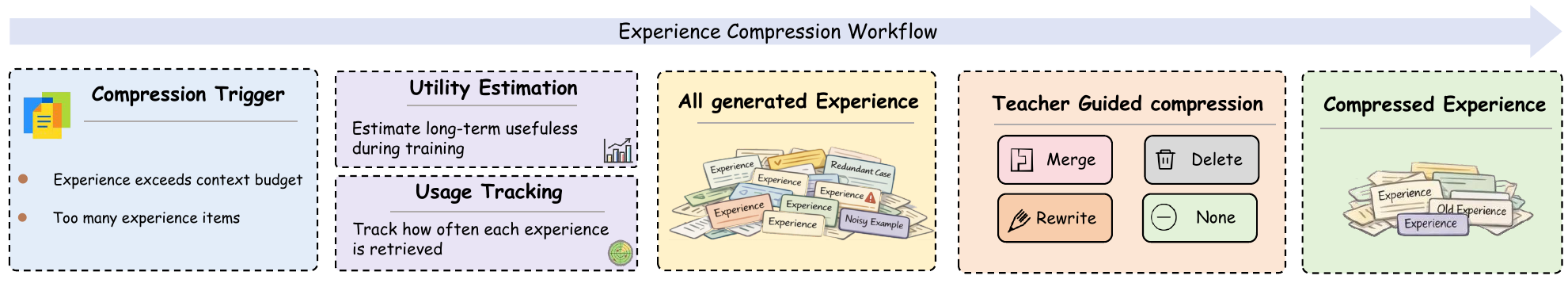}
\caption{Overview of the Experience Compression module in TED. When the experience  exceeds the context budget, TED estimates each experience’s utility and tracks its usage frequency. The teacher then compresses the experience by merging, rewriting, deleting, or retaining experiences, producing a compact system prompt that preserves high-utility knowledge for efficient, parameter-free iterative improvement.}
  \label{Fig3}
  \end{center}
\end{figure*}

\subsection{Experience Generation}

Based on the compressed reasoning trajectories, TED constructs and updates a reusable experience through teacher-driven critique. Given the student trajectories $\{\tau_i\}_{i=1}^{N}$, the teacher's valid trajectory $\tau_T$, and the ground-truth label $y$, the teacher model analyze differences between correct and incorrect reasoning paths and extract generalized experience.

\subsubsection{Teacher critique.}
The teacher jointly considers (i) multiple student reasoning trajectories, (ii) its own correct reasoning trajectory, and (iii) the ground-truth answer, and produces critiques that identify effective reasoning patterns, common failure modes, and corrective guidance.
Formally, we denote the critique process as
\begin{equation}
\mathcal{C} = \textsc{Critique}\big(\{\tau_i\}_{i=1}^{N},\tau_T,y\big),
\end{equation}
where $\mathcal{C}$ represents a set of candidate experience statements expressed in natural language.

\subsubsection{Experience update actions.}
TED maintains experience $E = \{e_j\}_{j=1}^{|E|}$, where each item $e$ encodes a reusable reasoning guideline or error pattern.
Instead of updating model parameters, TED updates $E$ by allowing the teacher to perform one of four discrete actions on the experience: 
\begin{itemize}
    \item \textbf{Add}: generate a new experience item and insert it into $E$;
    \item \textbf{Modify}: revise an existing experience item to improve correctness or generality;
    \item \textbf{Delete}: remove an obsolete or harmful experience item from $E$;
    \item \textbf{None}: take no action and keep $E$ unchanged.
\end{itemize}

\subsubsection{Positive-negative sample balance.}
To ensure stable experience generation, we control the balance between positive and negative student trajectories. 
For each input $x$, the $N$ student-generated trajectories are divided into correct and incorrect sets according to their final answers. 
We require the number of correct trajectories to be no smaller than the number of incorrect ones:
\begin{equation}
\big|\{\tau_i \mid \hat{y}(\tau_i)=y\}\big|
\;\ge\;
\big|\{\tau_i \mid \hat{y}(\tau_i)\neq y\}\big|.
\end{equation}

If no correct trajectory is produced, we keep only one negative trajectory to generate a critical experience. 

Through teacher critique, experience updates, and iterative refinement on balanced samples, TED builds an evolving experience that replaces parameter updates as the main mechanism for knowledge distillation.







\subsection{Experience Compression}

As training proceeds, the experience set $E$ may grow beyond the context limit and accumulate redundant or noisy items. TED therefore compresses $E$ to keep it compact while retaining useful information.

\subsubsection{Compression trigger.}
Let $B$ denote the maximum context budget measured in tokens, and let $\ell(e)$ be the serialized length of an experience item $e$.
Compression is triggered whenever
\begin{equation}
\begin{cases}
\sum_{e\in E}\ell(e) > B \\
|E| > B_{\text{item}}
\end{cases}
\end{equation}
where $B_{\text{item}}$ denotes the maximum number of experience items.

\subsubsection{Usage statistics and utility score.}
TED maintains usage statistics for each experience item across training.
Let $u_t(e)\in\mathbb{R}_{\ge 0}$ denote the accumulated usage frequency of item $e$ up to step $t$.
After processing sample $(x_t,y_t)$, the usage counter is updated as
\begin{equation}
u_t(e) = u_{t-1}(e) + \mathbb{I}\big[e \in \textsc{U}(E;x_t)\big],
\end{equation}
where $\textsc{U}(E;x_t)$ denotes the subset of experience items injected for input $x_t$, and $\mathbb{I}[\cdot]$ is the indicator function. During each forward inference, the model reports the IDs of the experience items it uses. We then count how many times each item is used.

We define a utility score $s_t(e)$, 
\begin{equation}
s_t(e) = \log(1 + u_t(e)).
\end{equation}

\subsubsection{Teacher-guided compression.}
At compression time, the teacher summarizes the experience into a smaller set $\hat{E}$. For a group of experience items, the teacher selects one of the following actions:
\begin{itemize}
    \item \textbf{Merge}: replace a set of redundant items with a single higher-level experience;
    \item \textbf{Rewrite}: rephrase an item to improve generality and applicability;
    \item \textbf{Delete}: remove obsolete, noisy, or harmful items;
    \item \textbf{None}: retain  unchanged.
\end{itemize}

\subsubsection{Utility-aware selection.}
During training, TED maintains usage statistics for each experience item. At compression time, the teacher performs utility-aware selection based on the accumulated usage frequency. Specifically, only the top-R most frequently used experiences are retained, while the remaining experiences are either merged with similar items or removed.

\section{Experiments}
To evaluate the performance of our TED, we present the implementation details, explain the experiments results, and offer a thorough analysis.

\subsection{Datasets}

We conducted experimental evaluations on multimodal mathematical reasoning benchmarks and visual logic benchmarks, including MathVision\cite{mathvision} and VisualPuzzles\cite{visualpuzzles}. In addition, we performed experiments on purely textual mathematical reasoning datasets. To ensure the reliability and robustness of our results, each problem was independently evaluated five times. We report the average score as Mean@5.
\begin{equation}
\mathrm{Mean@5}
= \frac{1}{N} \sum_{i=1}^{N}
\left(
\frac{1}{5} \sum_{j=1}^{5} s_{i,j}
\right)
\end{equation}
where $N$ denotes the total number of problems, and $s_{i,j}$ is the score obtained on the $i$-th problem in the $j$-th independent evaluation.

\subsection{Base Models}
In this paper, we adopt the Qwen3-VL \cite{Qwen3-VL} series as our student model and Kimi-K2.5\cite{kimi} as our teacher model, given the leading performance in the field of reasoning. We evaluate the effectiveness of our framework across multiple model sizes, including 8B and 235B, to demonstrate its ability to enhance models. To further explore the abilities of pure language models, we adopt Qwen3\cite{qwen3} series as student model and Kimi-k2.5 as teacher model.

\subsection{Hyper-parameters}
In order to ensure the stability of our experiment results, we standardized the hyper-parameters as follows. The temperature is fixed at 0.7, the top-p parameter is set to 1.0, and the max-token length is 32768. The max experience item is set to 15 and max context budget is 4000.

\subsection{Experimental Results}
Across all benchmarks, TED consistently improves over direct inference. Although fully trained knowledge distillation achieves the best overall performance, TED still obtains strong results with only a few hundred training samples and without updating model parameters. This makes TED a lightweight and practical alternative in low-data or resource-constrained settings. The results of all other baseline methods are obtained using their publicly available code.

\subsubsection{Results on the Multimodal Mathematical Reasoning}

We randomly sample 100 examples from MathVerse for training and evaluate on MathVision. The learning process runs for 3 epochs with a batch size of 5 and a group size of 5. For comparison, the Naive-KD baseline is trained on the full 3940-sample MathVerse set. All models are evaluated in thinking mode.

As shown in Table 1, TED consistently improves over direct inference despite using only a small number of training samples. For Qwen3-VL-8B, TED improves accuracy from 0.627 to \textbf{0.702}, and for Qwen3-VL-235B, from 0.746 to \textbf{0.762}. Although fully trained Naive-KD achieves higher absolute accuracy, TED remains competitive without parameter updates and with only 100 training examples. This suggests that distilled experiences stored in context can effectively transfer knowledge from teacher to student, especially for smaller models with limited capacity. Overall, TED offers a lightweight and data-efficient alternative to traditional KD, achieving substantial gains without expensive retraining.

Table~\ref{tab:data_size} further compares TED and Naive-KD under the same training data budgets using Qwen3-VL-8B as the student model. TED already performs well with 100 samples (0.702) and improves only slightly as more data is added, suggesting that it can learn useful experiences even in low-data settings. In contrast, Naive-KD depends more on larger training sets, improving from 0.629 with 100 samples to 0.764 with 3000 samples. These results suggest that TED works better in low-data or resource-constrained settings, while Naive-KD benefits more from larger datasets and eventually surpasses TED.

\begin{table}[htbp]
\centering
\caption{Results on the MathVision benchmark.}
\small
\label{tab:math}
\begin{tabular}{lccc}
\toprule
\textbf{Method} & \textbf{Train Set} & \textbf{Student} & \textbf{MathVision} \\
\midrule
Direct   & --        & Qwen3-VL-8B    & 0.627 \\
Direct   & --        & Qwen3-VL-235B  & 0.746 \\
\midrule
RAG \cite{ERP}     & --        & Qwen3-VL-8B    & 0.639 \\
RAG      & --        & Qwen3-VL-235B  & 0.751 \\
\midrule
Few-shot & --        & Qwen3-VL-8B    & 0.631 \\
Few-shot & --        & Qwen3-VL-235B  & 0.744 \\
\midrule
Naive-KD \cite{On-policy}       & MathVerse & Qwen3-VL-8B    & 0.729 \\
Naive-KD       & MathVerse & Qwen3-VL-235B  & 0.795 \\
\midrule 
Reflexion\cite{reflexion} & MathVerse & Qwen3-VL-8B & 0.662 \\
Reflexion & MathVerse & Qwen3-VL-235B & 0.751 \\
Memento \cite{Memento}  & MathVerse & Qwen3-VL-8B & 0.674 \\
Memento & MathVerse & Qwen3-VL-235B & 0.758 \\ 
MemCom \cite{Memcom} & MathVerse & Qwen3-VL-8B & 0.646 \\
MemCom & MathVerse & Qwen3-VL-235B & 0.741 \\
\midrule
TED    & MathVerse & Qwen3-VL-8B    & 0.702 \\
TED     & MathVerse & Qwen3-VL-235B  & 0.762 \\
\bottomrule
\end{tabular}
\end{table}

\begin{table}[htbp]
\centering
\caption{Comparison of different training data size}
\label{tab:data_size}
\begin{tabular}{lcccc}
\toprule
\textbf{Method} & \textbf{100} & \textbf{500} & \textbf{1000} & \textbf{3000}\\
\midrule
TED & 0.702 & 0.707 & 0.710 & 0.725\\
Naive-KD & 0.629 & 0.714 & 0.722 & 0.764\\
\bottomrule
\end{tabular}
\end{table}

\subsubsection{Results on the Multimodal Visual Logic}
To evaluate the generality of our method, we further conduct experiments on multimodal visual logic tasks under the same controlled setup. We randomly sample 100 training examples and evaluate on the VisualPuzzles benchmark, following the same learning setting as in the multimodal mathematical reasoning experiments.

As shown in Table 3, TED consistently improves over direct inference on this benchmark. For Qwen3-VL-8B, TED improves performance from 0.517 to \textbf{0.561}, and for Qwen3-VL-235B from 0.572 to \textbf{0.579}. Although fully trained Naive-KD remains slightly stronger in absolute performance, TED achieves competitive results without parameter updates and with only a small number of training samples. This shows that context-based knowledge transfer can be a practical and data-efficient option for multimodal reasoning tasks, especially when gradient-based retraining is not preferred or not possible.

\begin{table}[t]
\centering
\caption{Results on the VisualPuzzles benchmark.}
\small
\label{tab:logic}
\begin{tabular}{lccc}
\toprule
\textbf{Method} & \textbf{Train Set} & \textbf{Student} & \textbf{VisualPuzzles} \\
\midrule
Direct & --          & Qwen3-VL-8B    & 0.517 \\
Direct & --          & Qwen3-VL-235B  & 0.572 \\
\midrule
Naive-KD     & LogicVista  & Qwen3-VL-8B    & 0.566 \\
Naive-KD     & LogicVista  & Qwen3-VL-235B  & 0.582 \\
\midrule 
Reflexion\cite{reflexion} & LogicVista & Qwen3-VL-8B & 0.524 \\
Reflexion & LogicVista & Qwen3-VL-235B & 0.574 \\
\midrule
TED   & LogicVista  & Qwen3-VL-8B    & 0.561 \\
TED   & LogicVista  & Qwen3-VL-235B  & 0.579 \\
\bottomrule
\end{tabular}
\end{table}

\subsubsection{Results on the language only benchmark.} 
We randomly sample 100 training instances from DAPO-Math-17k \cite{DAPO} and evaluate on AIME25\cite{aime25}. The student models are from the Qwen3 series, with Kimi-K2.5 as the teacher. Training runs for 3 epochs with a batch size of 5, using a temperature of 0.7 and a group size of 5 during learning. All hyperparameters are the same as in the multimodal experiments, and Naive-KD is implemented under the same training schedule. All models are evaluated in thinking mode.

As shown in Table 4, TED consistently improves over direct inference in the language-only setting. For Qwen3-8B, accuracy increases from 0.673 to \textbf{0.733}, and for Qwen3-235B from 0.815 to \textbf{0.846}. This indicates that distilled experience remains effective in pure textual settings without any visual modality. Although Naive-KD achieves the highest absolute performance, TED remains competitive without gradient-based optimization, showing that contextual experience alone can provide meaningful gains in textual mathematical reasoning. The larger improvement on the smaller model further suggests that context-based knowledge transfer is especially beneficial for capacity-limited models.

\begin{table}[t]
\centering
\caption{Results on the textual AIME25 benchmark.}
\small
\label{tab:aime}
\begin{tabular}{lcccc}
\toprule
\textbf{Method} & \textbf{Train Set} & \textbf{Student} & \textbf{Teacher} & \textbf{AIME25} \\
\midrule
Direct & --         & Qwen3-8B    & --        & 0.673 \\
Direct & --         & Qwen3-235B  & --        & 0.815 \\
\midrule
Naive-KD     & DAPO-Math  & Qwen3-8B    & Kimi-K2.5 & 0.792 \\
Naive-KD     & DAPO-Math  & Qwen3-235B  & Kimi-K2.5 & 0.861 \\
\midrule
TED   & DAPO-Math  & Qwen3-8B    & Kimi-K2.5 & 0.733 \\
TED   & DAPO-Math  & Qwen3-235B  & Kimi-K2.5 & 0.846 \\
\bottomrule
\end{tabular}
\end{table}

\subsubsection{Analysis of training costs}

\begin{table}[t]
\centering
\caption{Training cost comparison on MathVerse under the same training budget (100 samples, $N=5$).}
\small
\begin{tabular}{lccc}
\toprule
\textbf{Method} & \textbf{GPU Hours} & \textbf{Monetary Cost (\$)} & \textbf{Reduction} \\
\midrule
Naive-KD & 576 & 288.0 & - \\
TED      & --  & 12.6  & 22.9$\times$ \\
\bottomrule
\end{tabular}
\label{tab:cost}
\end{table}

We compare the training costs of Naive-KD and TED on MathVerse under the same training budget, using 100 training samples and the same on-policy sampling setting ($N=5$). For Naive-KD, training is conducted on 8 NVIDIA A800 GPUs for 3 days, resulting in approximately 576 GPU-hours in total. Assuming a rental price of \$0.5 per GPU-hour, the overall training cost is about \$288.

By contrast, TED is training-free and only requires experience updates over the same 100 samples for 3 epochs. In our setting, the entire process finishes within 8 hours. During this procedure, the student model consumes approximately 21 million tokens, while the teacher model consumes about 6 million tokens. Based on the official pricing of Qwen3-VL and Kimi-K2.5, the total cost is only about \$12.6. 

Overall, TED reduces the training cost by more than one order of magnitude compared with Naive-KD, achieving a cost reduction of approximately $22.9\times$. These results demonstrate that TED is highly cost-effective, as it avoids expensive gradient-based optimization while still delivering strong performance.

\begin{figure*}[htbp]
  \centering

  \begin{subfigure}[b]{0.32\textwidth}
    \centering
    \includegraphics[width=\linewidth]{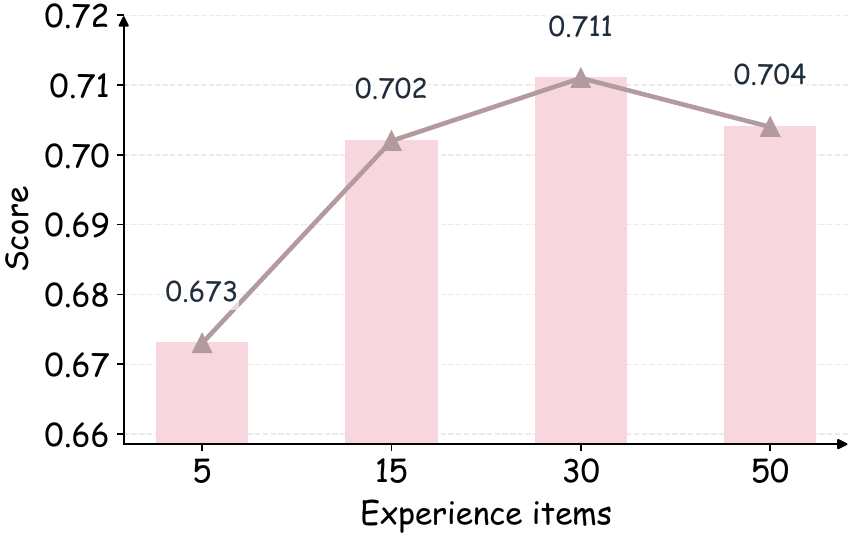}
    \caption{Experience Items Experiment}
    \label{fig:ablation_experience_items_paper}
  \end{subfigure}
  \hfill
  \begin{subfigure}[b]{0.32\textwidth}
    \centering
    \includegraphics[width=\linewidth]{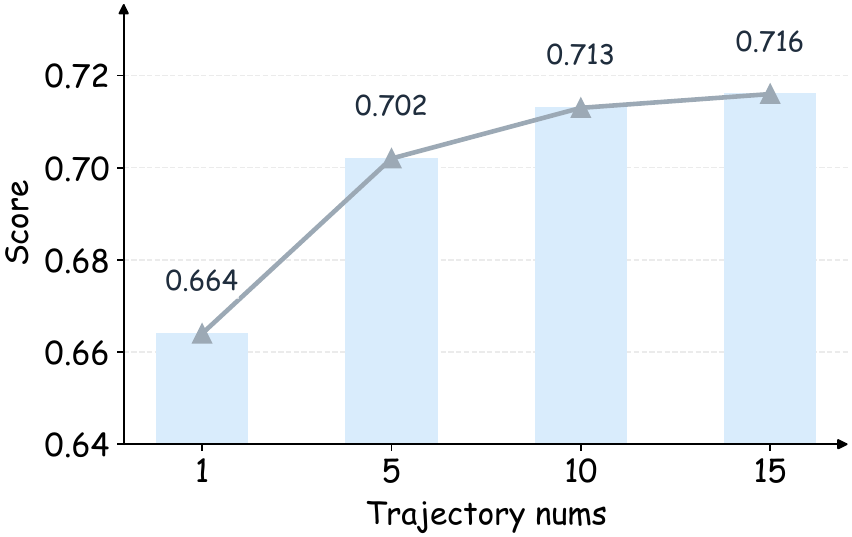}
    \caption{Trajectory Nums Experiment}
    \label{fig:ablation_trajectory_nums_paper}
  \end{subfigure}
  \hfill
  \begin{subfigure}[b]{0.32\textwidth}
    \centering
    \includegraphics[width=\linewidth]{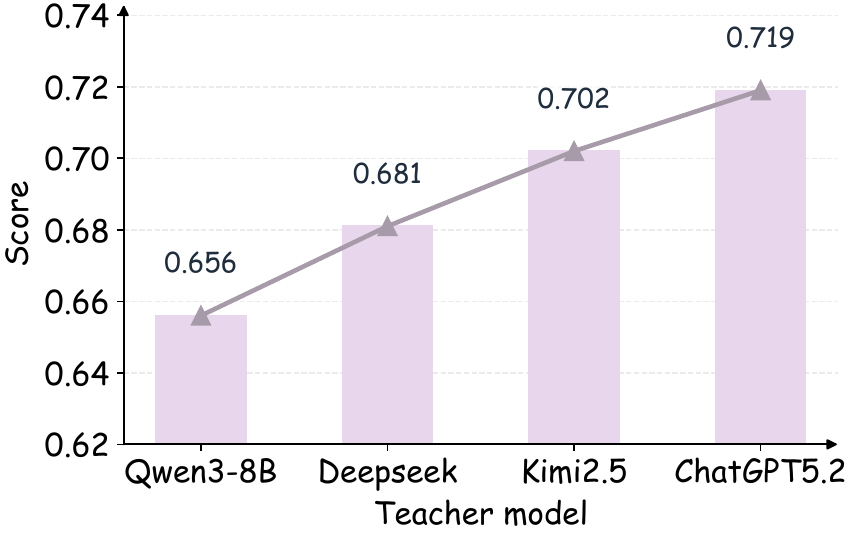}
    \caption{Teacher Models Experiment}
    \label{fig:ablation_teacher_models_paper}
  \end{subfigure}

  \caption{Hyperparameter ablation of TED on MathVision.
Performance is affected by the number of experience items, the number of sampled trajectories, and the choice of teacher model. TED performs best with a moderate experience size, more diverse trajectories, and stronger teachers.}
  \label{fig:hyperparameter_experiment}
\end{figure*}

\subsection{Ablation study}
To evaluate the effectiveness of our proposed method, we perform several ablation studies using Qwen3-VL-8B and MathVision benchmark. All other settings are kept unchanged.

\subsubsection{Effect of Teacher-Guided Experience}
\begin{table}[htbp]
\centering
\caption{Ablation Study of Teacher-Guided Experience}
\label{tab:teacher-guided}
\begin{tabular}{lccc}
\toprule
\textbf{Method} & \textbf{Train Set} & \textbf{Teacher} & \textbf{MathVision} \\
\midrule
Direct & -- & -- & 0.627 \\
Successful few-shot & MathVerse & - & 0.631 \\
TED   & MathVerse & Qwen3-VL-8B & 0.656 \\
TED   & MathVerse & Kimi-K2.5   & 0.702 \\
\bottomrule
\end{tabular}
\end{table}

As shown in Table~\ref{tab:teacher-guided}, \textit{successful few-shot} yields only a marginal improvement over direct inference (0.627 $\rightarrow$ 0.631), suggesting that simply retrieving successful examples is not sufficient to substantially enhance reasoning. In contrast, TED with the same student model as teacher (Qwen3-VL-8B) improves performance to 0.656, showing that the gain mainly comes from the \emph{experience learning mechanism}—iterative critique, refinement, and experience accumulation—rather than from few-shot examples alone. Using a stronger teacher (Kimi-K2.5) further boosts performance to 0.702, indicating that better feedback can produce higher-quality experience and lead to stronger generalization.
\subsubsection{Effect of Experience Compression}

\begin{table}[htbp]
\centering
\caption{Ablation Study of Experience Compression}
\label{tab:experience-compression}
\begin{tabular}{lcc}
\toprule
\textbf{Method}  & \textbf{Teacher} &\textbf{MathVision} \\
\midrule
Direct  & -- & 0.627 \\
\midrule
TED (Full) & Kimi-K2.5 & 0.702 \\
TED W/o Compression & Kimi-K2.5 & 0.594 \\
TED naive cut & Kimi-K2.5 & 0.648 \\
TED random choose & Kimi-K2.5 & 0.632 \\
\bottomrule
\end{tabular}
\end{table}

As shown in Table~\ref{tab:experience-compression}, experience compression is essential for effective experience learning. Without compression, performance drops sharply from 0.702 to 0.594, even below direct inference, showing that simply accumulating more experiences does not help and can instead hurt reasoning due to redundancy and noise. Naive cut (0.648) and random choose (0.632) both recover part of the performance by reducing experience length, but they remain clearly worse than full TED. In contrast, the full compression mechanism produces a compact and informative experience, achieving the best result of 0.702. These results verify the effectiveness of experience compression in making context-based distillation both stable and scalable.

\subsubsection{Cross-Modal Experience Transfer}
\begin{table}[htbp]
\centering
\caption{Cross-modal generalization of experience.}
\label{tab:cross_modal}
\small
\begin{tabular}{lcc}
\toprule
\textbf{Experience source} & \textbf{MathVision (Qwen-VL)} & \textbf{AIME25 (Qwen)} \\
\midrule
Direct & 0.627 & 0.673 \\
\midrule
MathVerse  & 0.702 & 0.686 \\
DAPO-Math  & 0.692 & 0.733 \\
\bottomrule
\end{tabular}
\end{table}

As shown in Table~\ref{tab:cross_modal}, TED demonstrates effective cross-modal transfer. Experience learned from multimodal MathVerse not only improves MathVision (0.627 $\rightarrow$ 0.702) but also brings gains on the text-only AIME25 benchmark (0.673 $\rightarrow$ 0.686). Similarly, experience learned from text-only DAPO-Math improves AIME25 more substantially (0.673 $\rightarrow$ 0.733) and also transfers well to multimodal MathVision (0.627 $\rightarrow$ 0.692). Although same-modality transfer yields the largest gains, the consistently positive cross-modal results show that TED captures transferable reasoning knowledge beyond modality-specific patterns.

\subsubsection{Detailed ablation study}

We further analyze three key factors of TED on MathVision, as shown in Figure~\ref{fig:hyperparameter_experiment}. 

First, increasing the number of experience items consistently improves performance from 0.673 to 0.711 as the number of experience items grows from 5 to 30, showing that richer experience provides more useful reasoning guidance. When the number of experience items is further expanded to 50, performance slightly drops to 0.704, suggesting that overly long experience introduces redundancy and reduces context efficiency. Therefore, to balance cost, context length, and overall performance, we choose 15 experience items as the default setting.

Second, increasing the number of sampled trajectories steadily improves performance from 0.664 to 0.716. Even a single trajectory already brings gains, indicating that teacher critique alone can refine reasoning to some extent. However, multiple trajectories provide more diverse correct and incorrect reasoning paths, enabling the teacher to extract more generalizable experience knowledge.

Third, TED also benefits from stronger teacher models. Using the student itself as teacher already achieves 0.656, while stronger teachers further improve performance from 0.681 (DeepSeek) to 0.702 (Kimi2.5) and 0.719 (ChatGPT5.2\cite{chatgpt}). This trend shows that better teachers provide higher-quality supervision for experience extraction, leading to more informative and transferable experience.

Overall, these results show that TED is robust to different design choices, and its performance is jointly determined by experience capacity, trajectory diversity, and teacher quality.

\section{Conclusion}
In this paper, we present TED, a parameter-free and context-based knowledge distillation framework that transfers knowledge through contextual experience accumulation rather than gradient-based optimization. By distilling teacher-guided reasoning trajectories into compact and reusable experiences, TED enables student models to continually improve without updating their parameters. 

Extensive experiments on multimodal reasoning and textual mathematical benchmarks demonstrate that TED consistently improves model performance in low-data settings while requiring only a small number of training samples. Across both multimodal and textual tasks, TED provides a strong performance-cost trade-off and achieves results competitive with conventional parameter-based distillation, while reducing training cost by more than 20× and avoiding parameter optimization. The proposed framework also shows encouraging generalization across different modalities and model scales. 

At the same time, TED is not intended to replace gradient-based distillation in large-scale settings with abundant data and sufficient resources, where full parameter updates can still achieve stronger performance. These results suggest that substantial knowledge transfer can be achieved through contextual experience injection, making TED a lightweight, data-efficient, and practical alternative for scenarios where conventional retraining is costly or infeasible.

\bibliographystyle{ACM-Reference-Format}
\bibliography{sample-base}

\clearpage

\onecolumn
\appendix

\section{Limitation and discussion}
Despite the promising results, TED also has several limitations. The proposed framework is particularly suitable for scenarios with limited training data, restricted computational resources, or black-box APIs, where gradient-based optimization is impractical or unavailable. By operating entirely in the context space, TED avoids parameter updates and significantly reduces training cost. However, in settings with large-scale datasets and sufficient computational resources, gradient-based distillation or fine-tuning methods are still likely to achieve stronger performance, as they can fully update model parameters and exploit larger training signals.

From a qualitative perspective, the effectiveness of TED may stem from its ability to reduce the reasoning search space for smaller models. When solving complex problems, small models often face a large and inefficient reasoning search space during the thinking process. The trajectory-based framework of TED iteratively collects reasoning paths and leverages teacher critiques together with positive and negative trajectory comparisons, which gradually constrain the exploration space and guide the model toward more effective reasoning strategies. As a result, even self-distillation or relatively weak teacher models can still provide useful signals that improve performance. When stronger teacher models are used, their reasoning trajectories tend to be more accurate and efficient, leading to higher-quality experience extraction and larger performance gains. In practice, the distilled experiences often function as high-level reasoning patterns or thinking templates, which help smaller models adopt more structured and effective problem-solving strategies.

\section{Details of prompts}

This section reports the exact prompts used in our experiments.

\begin{promptbox}{Inference with experience}
Please solve the problem in the figure:
{problem} 

When solving problems, you MUST first carefully read and understand the helpful instructions and experiences:
{experiences}

Final answer should be start with Answer
for example:
Answer: A/B/C/D
\end{promptbox}

\begin{promptbox}{Inference of teacher}
Please solve the problem:
{problem}

Final answer should be start with <Answer>

for example:
<Answer:> A/B/C/D
\end{promptbox}

\begin{promptbox}{Prompt of teacher critique}
You are given:
(1) a problem,
(2) multiple solution trajectories generated by a student network,
(3) one or more trajectories generated by a teacher network.

The objective is to extract generalizable reasoning experiences from the teacher trajectories that can guide and correct the student network in future attempts on structurally similar problems.

Student trajectories are labeled with binary rewards:
- POSITIVE (reward = 1): successful solutions
- NEGATIVE (reward = 0): failed solutions.

You must perform a structured comparative analysis before updating experiences.

1. Comparative Trajectory Analysis

Teacher Trajectories:
- Identify the key strategic decisions and pivotal reasoning steps.
- Analyze how critical decision points, ambiguities, and potential pitfalls are resolved.
- Distill recurring reasoning patterns that contribute to correctness.

Student Trajectories:

Positive Trajectories (reward = 1):
- Identify strategies that directly contributed to success.
- Analyze alignment with the teacher’s reasoning patterns.
- Determine which reasoning components are transferable.

Negative Trajectories (reward = 0):
- Identify divergence points from successful reasoning paths.
- Categorize error patterns, such as:
  • misapplied principles,
  • overlooked constraints,
  • incorrect assumptions,
  • premature simplifications,
  • local optimization traps.
- Extract partially correct but incomplete reasoning components.

Cross Comparison:
- Identify decisive differences between positive and negative trajectories.
- Determine reasoning strategies consistently applied by the teacher but absent in failed student attempts.

2. Updating the Experience Set

Teacher trajectories may include both correct and incorrect reasoning paths.
Only extract experiences that reliably promote correct strategic reasoning.

You may perform one of the following operations:
- "modify": refine an existing experience to improve clarity or correctness.
- "add": introduce a new generalizable experience.
- "delete": remove an incorrect or misleading experience.
- "nan": make no updates if the experience set is already sufficient.

3. Experience Formulation Requirements

Each experience must:
- Begin with a concise description of the general problem context.
- Emphasize strategic reasoning patterns rather than specific computations.
- Highlight reusable decision points applicable across similar tasks.
- Avoid referencing specific numeric values or problem-dependent details.
\end{promptbox}

\begin{promptbox}{Prompt of Trajectory compression}
You are given a problem, and the following rollouts to solve the given problem. Please summarize the trajectory step-by-step:

For each step, describe **what action is being taken**. Only the abstract experiences.

<problem>
{problem}
<problem>

<rollouts>
{rollouts}
</rollouts>

only return the summary of each step, e.g.,
1. what happened in the first step and the core outcomes
2. what happened in the second step and the core outcomes
\end{promptbox}

\begin{promptbox}{Prompt of Experience compression}
An agent system maintains a set of reasoning experiences.
Currently, there are some  experiences, which introduces redundancy and weakens strategic clarity.

Your objective is to perform experience compression under the FREE framework by consolidating overlapping reasoning patterns and removing duplication. The final number of experiences must not exceed 15.

Each resulting experience must satisfy the following criteria:

1. It must express a clear and generalizable strategic lesson, within 32 words.
2. It must begin with a concise general background context.
3. It must focus on reasoning strategies rather than specific computations.
4. It must emphasize transferable decision points applicable to similar problems.
5. It must avoid semantic overlap with other retained experiences.

You are provided with:
{experiences}

Compression Procedure:

1. Redundancy Analysis  
   - Identify experiences expressing similar strategic principles.
   - Detect overlap in decision logic, structural reasoning, or error prevention themes.
   - Group experiences that differ superficially but share core reasoning patterns.

2. Strategic Abstraction  
   - Generalize grouped experiences into a higher-level strategic principle.
   - Remove problem-specific language.
   - Preserve critical decision-point structure.

3. Compression Operations  

You may use the following update operations:

- "modify": refine an existing experience to improve abstraction and generality.
- "merge": combine multiple similar experiences into one more general and strategically expressive experience.  
  (Merge is the primary mechanism for reducing count.)
\end{promptbox}

\section{Some examples of experience}
\begin{promptbox}{Some examples of experience}
\textbf{E1}: For signed quantities with unknown signs, introduce sign variables, enumerate cases systematically, and optimize. Extrema often emerge when sign groups oppose each other.

\textbf{E2}: For optimization, push variables to constraints to find bounds, verify attainability. For maximin problems, identify the most restrictive objective to bound the optimum, then ensure others can meet it.

\textbf{E3}:For random walk hitting probabilities: model as renewal process, derive first-step recurrence, encode as generating function, analyze dominant singularity for limiting behavior.

\textbf{E4}: Geometric algebra: Use mass points with scaled vertex masses for ratios. Or translate to coordinates/vectors, exploit symmetry, compute via formulas, and identify linear relations.

\textbf{E5}:Geometry properties: For angle bisectors, use excenter relationships and half-angle formulas. For circle tangents, exploit perpendicularity to form right triangles and apply trigonometric ratios.

......
\end{promptbox}

\clearpage

\section{Details of algorithm}

\begin{algorithm}[htbp]
\caption{Reasoning Trajectory Generation in TED}
\label{alg:ted_traj_gen}
\KwIn{input--label pair $(x,y)$, student model $S$, teacher model $T$, experience $E$, trajectory number $N$}
\KwOut{condensed student trajectories $\{\tau_i\}_{i=1}^{N}$, teacher trajectory $\tau_T$, teacher-validity flag}

Construct the prompted context $p(x;E)=[p_{\mathrm{sys}};E;x]$\;

Sample $N$ student raw trajectories in parallel:
$\{\tilde{\tau}_i\}_{i=1}^{N} \sim S(\cdot \mid p(x;E))$\;

Generate the teacher raw trajectory:
$\tilde{\tau}_T \sim T(\cdot \mid x)$\;

\For{$i \leftarrow 1$ \KwTo $N$}{
    $\tau_i \leftarrow \mathrm{Condense}(\tilde{\tau}_i)$\;
}
$\tau_T \leftarrow \mathrm{Condense}(\tilde{\tau}_T)$\;

Enforce the structured format:
\texttt{Premises} $\rightarrow$ \texttt{Step 1} $\rightarrow$ \texttt{Step 2} $\rightarrow \cdots \rightarrow$ \texttt{Conclusion}\;

\eIf{$\hat{y}(\tau_T)=y$}{
    mark $\tau_T$ as valid\;
}{
    mark $\tau_T$ as invalid and treat this sample as a negative case for later critique\;
}

\Return{$\{\tau_i\}_{i=1}^{N}, \tau_T$}\;
\end{algorithm}

\begin{algorithm}[htbp]
\caption{Experience Generation in TED}
\label{alg:ted_exp_gen}
\KwIn{student trajectories $\{\tau_i\}_{i=1}^{N}$, teacher trajectory $\tau_T$, ground-truth label $y$, experience $E$}
\KwOut{updated experience $E$}

Partition student trajectories into
$\mathcal{T}^{+}=\{\tau_i \mid \hat{y}(\tau_i)=y\}$ and
$\mathcal{T}^{-}=\{\tau_i \mid \hat{y}(\tau_i)\neq y\}$\;

\If{$|\mathcal{T}^{+}| < |\mathcal{T}^{-}|$}{
    down-sample $\mathcal{T}^{-}$ until
    $|\mathcal{T}^{+}| \ge |\mathcal{T}^{-}|$\;
}

\If{$|\mathcal{T}^{+}| = 0$}{
    keep only one negative trajectory in $\mathcal{T}^{-}$\;
}

Generate teacher critique:
$C=\mathrm{Critique}(\{\tau_i\}_{i=1}^{N}, \tau_T, y)$\;

Teacher selects one action from
$\{\mathrm{Add}, \mathrm{Modify}, \mathrm{Delete}, \mathrm{None}\}$\;

\Switch{selected action}{
\Case{$\mathrm{Add}$}{
    insert a new experience item into $E$\;
}
\Case{$\mathrm{Modify}$}{
    revise an existing experience item in $E$\;
}
\Case{$\mathrm{Delete}$}{
    remove an obsolete or harmful experience item from $E$\;
}
\Case{$\mathrm{None}$}{
    keep $E$ unchanged\;
}
}

\Return{$E$}\;
\end{algorithm}

\begin{algorithm}[t]
\caption{Experience Compression in TED}
\label{alg:ted_exp_comp}
\KwIn{experience $E=\{e_j\}_{j=1}^{|E|}$, context budget $B$, item budget $B_{\mathrm{item}}$, step $t$, current sample $(x_t,y_t)$}
\KwOut{compressed experience $\hat{E}$}

\ForEach{$e \in U(E;x_t)$}{
    $u_t(e) \leftarrow u_{t-1}(e) + \mathbb{I}[e \in U(E;x_t)]$\;
    $s_t(e) \leftarrow \log(1 + u_t(e))$\;
}

\If{$\sum_{e\in E}\ell(e) > B$ \textbf{or} $|E| > B_{\mathrm{item}}$}{
    retain only the top-$R$ most frequently used / highest-utility experiences\;

    Teacher summarizes $E$ into a smaller set $\hat{E}$\;

    \ForEach{candidate item or item group in $E$}{
        select one action from
        $\{\mathrm{Merge}, \mathrm{Rewrite}, \mathrm{Delete}, \mathrm{None}\}$\;

        \Switch{selected action}{
        \Case{$\mathrm{Merge}$}{
            replace redundant items with one higher-level experience\;
        }
        \Case{$\mathrm{Rewrite}$}{
            rephrase an item to improve generality and applicability\;
        }
        \Case{$\mathrm{Delete}$}{
            remove obsolete, noisy, or harmful items\;
        }
        \Case{$\mathrm{None}$}{
            retain the item unchanged\;
        }
        }
    }
}
\Else{
    $\hat{E} \leftarrow E$\;
}

\Return{$\hat{E}$}\;
\end{algorithm}









\end{document}